\documentclass{article}

     \usepackage[nonatbib]{neurips_2020}

\usepackage[utf8]{inputenc} 
\usepackage[T1]{fontenc}    
\usepackage{hyperref}       
\usepackage{url}            
\usepackage{booktabs}       
\usepackage{amsfonts}       
\usepackage{nicefrac}       
\usepackage{microtype}      
\usepackage{mathrsfs}
\usepackage{graphicx}
\usepackage{subfigure}
\usepackage{amsmath}
\usepackage{booktabs}

\usepackage[noend]{algpseudocode}

\usepackage{algorithmicx,algorithm}

\title{Transport based Graph Kernels }

\author{
  Kai Ma\\
  \texttt{kaim@nuaa.edu.cn}
  \And
  Peng Wan \\
  \texttt{wanpengwx@163.com}
  \And
  Daoqiang Zhang\thanks{Corresponding author} \\
  \texttt{dqzhang@nuaa.edu.cn}
  \\
  Department of Computer Science and Technology\\
  MIIT Key Laboratory of Pattern Analysis and Machine Intelligence\\
  Nanjing University of Aeronautics and Astronautics\\
  Jiangsu Province, China, 210016 \\
}

\begin{document}

\maketitle

\begin{abstract}
Graph kernel is a powerful tool measuring the similarity between graphs. Most of the existing graph kernels focused on node labels or attributes and ignored graph hierarchical structure information. In order to effectively utilize graph hierarchical structure information, we propose pyramid graph kernel based on optimal transport (OT). Each graph is embedded into hierarchical structures of the pyramid. Then, the OT distance is utilized to measure the similarity between graphs in hierarchical structures. We also utilize the OT distance to measure the similarity between subgraphs and propose subgraph kernel based on OT.  The positive semidefinite (p.s.d) of graph kernels based on optimal transport distance is not necessarily possible. We further propose regularized graph kernel based on OT where we add the kernel regularization to the original optimal transport distance to obtain p.s.d kernel matrix. We evaluate the proposed graph kernels on several benchmark classification tasks and compare their performance with the existing state-of-the-art graph kernels. In most cases, our proposed graph kernel algorithms outperform the competing methods.

\end{abstract}

\section{Introduction}
Graph structure data, including biological networks~\cite{YouGraph2018,JustinNeural2017}, social networks~\cite{Xue2016VoteTrust,Bhagat2009Class} and citation networks~\cite{Citeseer1998,2000Automating,Leydesdorff2018Betweenness}, widely exist in the real world. One crucial problem for graph structure data in machine learning is measuring the similarities between graphs with attribute information and connectivity structure. Graph kernels~\cite{2010Graph} have shown high success in the similarity measurements and powerful performance on all kinds of classification tasks~\cite{morris2016faster,yanardag2015deep}.

A variety of graph kernels have been proposed. They utilized the different strategies to measure the similarities between graphs. The strategies include shortest path~\cite{Borgwardt2006Shortest}, graphlets~\cite{Nino2009Efficient}, subtrees~\cite{Thomas2003}, cycles~\cite{Thomas2004}, and random walks~\cite{kashima2003marginalized,2018RetGK}. These kernels used special structures (e.g., shortest path, subtree) to represent the graphs and then calculated the similarities based on the special structures. The special structure of the graph usually reflected graph special characteristic. For example, the shortest path in shortest-path kernels reflected the path characteristic of the graph, then the kernel is calculated on the shortest path between two nodes in graph. The graph kernels based on special structures usually focused on the local properties and ignored the global structure information.

Graph and subgraph isomorphism, checking whether two graphs or subgraphs have different size~\cite{Shervashidze2011Weisfeiler} in spatial structure, could be used to measure the topological similarity of global structure between two graphs (or subgraphs)~\cite{2012Subgraph}. Unfortunately, graph or subgraph isomorphisms is NP problems~\cite{AM1979}. The graph kernels based on graph or subgraph isomorphism are hardly calculated in polynomial time. To overcome the isomorphism problem in graph or subgraph, some graph or subgraph kernels based on node or edge attributes were proposed~\cite{2012Subgraph,morris2016faster,feragen2013scalable}. For example, Kriege and Mutzel proposed the subgraph matching kernels~\cite{2012Subgraph}, which could rate the mappings of subgraph by comparing edge and node attributes using the related scoring scheme. However, these existing graph or subgraph kernels based on attributes still ignored the global structure information of the graph.

Recently, optimal transport (OT) theory has attracted broad attention in image processing~\cite{ferradans2014regularized}, computer vision~\cite{Fitschen2017Optimal} and neural network~\cite{2017WassersteinGAN}. OT distance~\cite{Villani2008} is designed to measure the similarity between two probability distributions when the ground distance has been given. OT distance could capture the geometric information of the underlying probability space.
Graph structure data could be represented as probability distributions with the related graph or network embedding methods. Hence, OT could be used to measure the similarity between graphs and capture graph global structure information. Nikolentzos et al.~\cite{Nikolentzos2017Matching} utilized the eigen decomposition of adjacency matrix to represent the graphs as bag-of-vectors and then used OT distance to measure the similarity between two graphs. Their method used OT to capture the graph global structure and measured the graph similarity.
In their research, they also introduced pyramid match graph kernel where the graph similarity was measured based on the intersections of multi-resolution histograms. Their methods ignored the graph global structure information in hierarchical structures. Togninalli et al.~\cite{NIPS2019_8872} used Weisfeiler-Lehman scheme to produce the node label sequences and then utilized the OT distance to measure the similarity between label sequences. This method ignored the structural information of graphs and the positive semidefinite (p.s.d) of graph kernels were not necessarily possible.

In this paper, we propose pyramid graph kernel based on OT (PG-OT kernel) which measure the similarities between graph global information in hierarchical structures. We embed the graphs into the pyramid structure and then utilize the OT distance to measure the similarities between graphs in each hierarchical structure. We also present a subgraph kernel based on OT (SG-OT kernel) for addressing the problem that the subgraph kernels based on attributes ignore the global structure information. For the reason that the graph kernels based on OT distance are usually not positive semidefinite (p.s.d), we further propose the regularized graph kernel based on OT (RG-OT kernel), which could obtain the p.s.d kernel matrix.

The rest of this paper is organized as follows. We first introduce the preliminary for our research. Then, we introduce the optimal transport distance on graphs and our proposed graph kernels based on optimal transport. At last, we evaluate our proposed methods and compare them with the start-of-the-art graph kernels.


\section{Preliminary}
In this section, we first introduce some notations related to graph and basic concepts about graph kernels. Then, we introduce a graph representation called bags-of-vectors. Subsequently, we introduce the content related to the pyramid match and subgraph match and the problems existing in them. At last, we introduce a distance metric called optimal transport distance.

\subsection{Notations}
$G=(V,E)$ is an undirected graph which consists of a set $V$ of nodes and a set $E$ of edges. $A$ is the adjacency matrix of graph $G$. $A_{ij}$ is the element in the $i^{th}$ row and $j^{th}$ column of $A$. $A_{ij}$ is represented as follows:
\\
\begin{equation}
A_{ij}=\left\{
\begin{aligned}
&1 , ~if (v_i,v_j)\in E, \\
&0 , ~otherwise.
\end{aligned}
\right.
\end{equation}
\\
where $(v_i,v_j)$ is the edge between node $v_i$ and node $v_j$, $v_i,v_j\in V$.

 $\langle\cdot\rangle$ is the Frobenius dot product. $X$ and $Y$ are two matrices, $X,Y\in\mathbb{R}^{m\times n}$, $\langle X,Y\rangle=tr(X^{T}Y)=\sum^{m}_{i=1}\sum^{n}_{j=1}X_{ij}Y_{ij}$. The simplex is denoted as $\sum_d :=\{x\in\mathbb{R}^{d}_{+} : x^{T}\textbf{1}_d=1\}$, where $\textbf{1}_d$ is the $d$-dimensional vector of ones.
 $\phi$ is a non-linear transformation, and the Euclidean distance based on $\phi$ could be defined as
 \\
\begin{equation}
    D_{\phi}(\textbf{u}, \textbf{v})=\|\phi{(\textbf{u})}-\phi{(\textbf{v})}\|_2
\end{equation}
\\
$D_{\phi}$ satisfies the definition of pseudo-metric in the original input space~\cite{Kedem2012Non}. $\textbf{u}$ and $\textbf{v}$ are vectors.

\subsection{Graph kernels}
Graph kernels are a class of kernel functions measuring the similarities between graphs. There is a map $\phi$, which could implicitly embed the original graph data set $\mathcal{G}$ into a Hilbert space $\mathcal{H}$, $\phi$: $\mathcal{G}$ $\rightarrow$ $\mathcal{H}$. In $\mathcal{G}$, graph kernel $\mathcal{K}$: $\mathcal{G}$ $\times$ $\mathcal{G}$ $\rightarrow$ $\mathbb{R}$ is a function associated with $\mathcal{H}$, given two graphs \emph{G$_1$} and \emph{G$_2$}, \emph{G$_1$}, \emph{G$_2$} $\in$ $\mathcal{G}$, graph kernel $\mathcal{K}$ is interpreted as a dot product in the high dimensional space $\mathcal{H}$, $\mathcal{K}\left(G_1,G_2\right)=\langle\phi\left(G_1\right),\phi\left(G_2\right)\rangle_\mathcal{H}$. The kernel function $\mathcal{K}$ is symmetric, i.e., $\mathcal{K}(G_1,G_2)=\mathcal{K}(G_2,G_1)$. If the Gram matrix $\textbf{K}\in\mathbb{R}^{N\times N}$ defined by $\textbf{K}(i,j)=\mathcal{K}(G_i,G_j)$ for any graph $\{G_i\}^{N}_{i=1}$ is \emph{positive semidefinite (p.s.d)}, then $\mathcal{K}$ is p.s.d.

\subsection{Bags-of-vectors}
Bags-of-vectors in graphs are inspired by natural language processing and image recognition where documents and images are described by bags-of-words and bags-of-features. Given a graph $G=(V,E)$, $V$ is a set of nodes and $E$ is a set of edges. Graph $G$ could be represented as bags-of-vectors like $\{u_1,u_2,...,u_n\}$, $u_i\in\mathbb{R}^d$, $i=1,2,...,n$, $u_i$ is a vector representation of node $v_i, v_i\in V$. Node vector representations are that nodes in graph are embedded into the vector space with the related embedding algorithms, e.g, eigenvalue decomposition~\cite{Nikolentzos2017Matching}, deep walk~\cite{perozzi2014deepwalk}, and Weisfeiler-Lehman scheme~\cite{NIPS2019_8872}.

\subsection{Pyramid match}
Pyramid match (PM) was proposed by Grauman and Darrell~\cite{grauman2007the} to learn efficient feature sets from a single sample. The unordered feature sets in a sample were projected into multi-resolution histograms. Then weighted histogram intersections were calculated for designing pyramid match kernel to measure the similarity between two feature sets. Su et al~\cite{Su2016A} proposed the vocabulary-guided pyramid match kernel for measuring the similarity between the node vector representations of two graphs. These two kernels used the labels and neighbors of nodes to represent the node features and ignored the graph global properties. To take the graph global properties into the pyramid match, Nikolentzos et al. represented each graph as a bags-of-vectors where they utilized the eigenvectors of graph adjacency matrices to generate the node embeddings~\cite{Nikolentzos2017Matching}. They projected the bags-of-vectors into multi-resolution histograms, and then computed the weighted histogram intersections for designing PM graph kernel to measure the similarity between two sets of vectors. PM graph kernel based on the eigenvectors of adjacency still calculated matrices histogram intersections and ignored the geometric information in each resolution histogram.

\subsection{Subgraph match}
Graph $G$ consists of multiple subgraphs and each subgraph uncovers the local geometric information. Graph kernels based on subgraph structures~\cite{bai2016fast,Tsuda2004Learning,Costa2010Fast} are widely used to compare graph similarities in numerous domains, e.g., cheminformatics, bioinformatics and brain networks. The representative kernel among them is subgraph matching kernel~\cite{kriege2012subgraph} which could measure the similarity between attributed graphs. This kernel could rate the mappings of subgraph by comparing edge and vertex attributes using the related scoring scheme. Most subgraph kernels fall into two main categories: (1) subgraph isomorphism, which is the analogue of graph isomorphism; (2)node or edge attribute information. The subgraph isomorphism problem is NP-complete, which could not be calculated by a polynomial-time algorithm~\cite{AM1979}. The subgraph kernels using node or edge attribute information potentially ignore the local geometric information.
\\ \\
\textbf{Theorem 1} \emph{Computing the subgraph isomorphism is NP-hard.}

\subsection{Optimal Transport Distance}
Optimal Transport (OT) is originally proposed to investigate the translocation of masses~\cite{Kantorovitch1958On}. In mathematics, OT is utilized to determine the difference between two probability distribution after giving ground metric. The distance induced by OT is also called Earth Mover's distance or Wasserstein distance~\cite{Peng2018Label}. Recently, OT is becoming extremely popular in image processing, computer vision and graph similarity measurement, such as  image reconstruction~\cite{I2017Tomographic,zheng2019guided}, image segmentation~\cite{Rabin2015Convex} and Wasserstein Weisfeiler-Lehman graph kernels~\cite{NIPS2019_8872}.

The optimal transport distances are very powerful in measuring two probability distributions, especially the probability space has geometrical structures. For $r$ and $c$ are two probability distributions from the simplex $\sum_n$ and $\sum_m$, we regard $\Gamma(r,c)$ as the set of all transportation plans of $r$ and $c$, $\Gamma$ is a $n\times m$ matrices.\\
\begin{equation}
\Gamma(r,c) :=\{P\in\mathbb{R}^{n\times m}_{+}| P\textbf{1}_m=r, P^T\textbf{1}_n=c\}
\end{equation}
\\
Given a $n\times m$ cost matrix $M$ defined on the ground distance $d(x,y)$, $x\in r, y\in c$, the total cost of mapping r to c with a transportation matrix $P$ could be represented as $\langle P,M\rangle$. The optimal transport distance could be defined as:
\\
 \begin{equation}
 d_{OT}(r,c) := min_{P\in\Gamma(r,c)}\langle P,M\rangle
 \label{OT_distance}
 \end{equation}
 \\
 The optimal transport distance is also called Wasserstein distance, hence formula~$(\ref{OT_distance})$ could also be written as $L^{P}$-Wasserstein distance for $p\in[1,\infty)$
 \begin{equation}
 W_{P}(r,c) := (inf_{\gamma\in\Gamma(r,c)}\int_{M\times M}d(x,y)^{p}d\gamma(x,y))^{\frac{1}{p}}
 \label{Wasserstein_distance}
 \end{equation}
However, most of the existing graph kernels based on OT represented each node in the graph with label sequences and discarded the graph geometrical structure information.

\begin{figure}
\centering
\includegraphics[width=8cm,height=4.2cm]{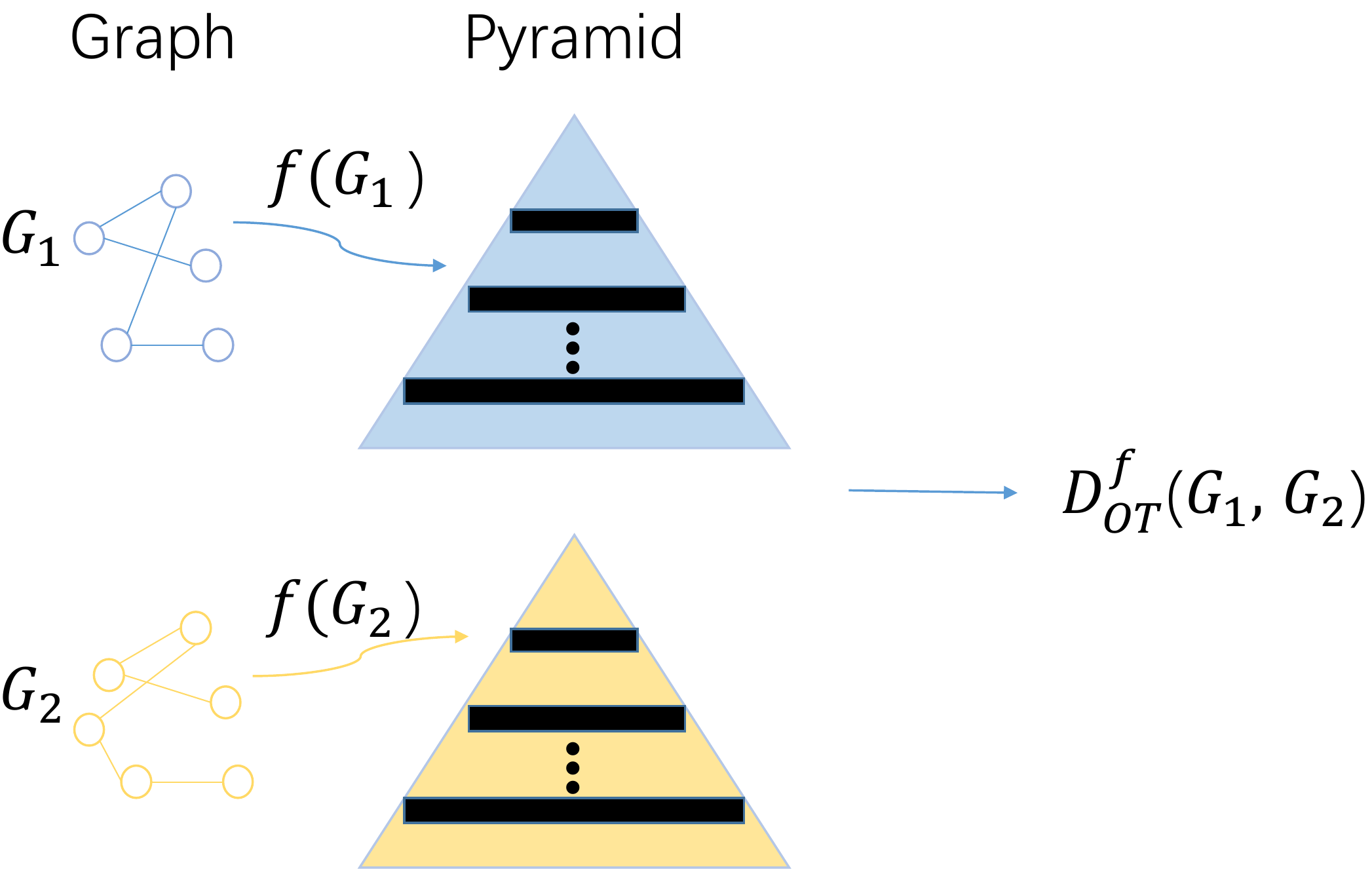}
\caption{Optimal transport distance based on Pyramid structure}
\label{Pyramid_graph_OT}
\end{figure}

\section{Optimal Transport Distance on Graphs}
The optimal transport distance is good at measuring the differences between two probability distributions. Hence, before applying the optimal transport distance to graph measurement, the graphs need to be transformed into probability distributions. Briefly speaking, the graph stricture data need to be transformed into vector data, i.e., graph vectorial representations. We usually adopt graph embedding strategies to obtain graph vectorial representations.
\\ \\
\textbf{Definition 1.} (Graph Embedding Strategy) Given a graph $G$, the graph embedding strategy could be regarded as a function $f$ transforming $G$ into vector space, like $f: G\to\mathbb{R}^{m\times n}, f(G)=X_{G}, X_{G}\in\mathbb{R}^{m\times n}$. $X_{G}$ is the graph vectorial representation. If $m=|V|, n=d$, i.e., $m$ is the number of nodes in the graph, $d$ is the vector dimension, $X_{G}$ is the vectorial representations of nodes in the graph. For node $v_{i}\in V$, the $i^{th}$ row of $X_{G}$ is the node embedding of $v_i$ and its dimension is $d$.
\\ \\
\textbf{Definition 2.} (Graph OT Distance) Given two graphs $G_1=(V_1,E_1)$ and $G_2=(V_2,E_2)$, a graph embedding strategy $f: G\to\mathbb{R}^{m\times n}$ and a ground distance $d$, the graph OT distance between $G_1$ and $G_2$ is defined as
\\
 \begin{equation}
    D^{f}_{OT}(G_1,G_2)=d_{OT}(f(G_1),f(G_2))
 \end{equation}

\subsection{Eigen-decomposition strategy}
Graph eigen-decomposition could be regarded as a graph or node embedding strategy. Given a graph $G$, $A$ is its adjacency matrix. The eigenvalue decomposition of $A$ is $A=U\Lambda U^T$, $Q$ is the orthogonal matrix of the eigenvectors of $A$, $\Lambda$ is diagonal matrix consisting of eigenvalues of $A$. The $i^{th}$ row $u_i$ of $U$ is the embedding of node $v_i$ in graph $G$. Because, $A$ is a real and symmetric matrix, its eigenvalues are still real. According~\cite{2008Social}, the $i^{th}$ component of the eigenvector of $A$ offers us the eigenvector centrality score of node $v_i$ in graph $G$. In our work, we embed all nodes in graph $G$ into $d$-dimensional vector space $\mathbb{R}^d$ and obtain the $d$-dimensional vectorial representations of all nodes with the eigenvectors of the $d$ largest in magnitude eigenvalues. Given a graph $G$, $f_{E}(G)$ denotes the eigen-decomposition embedding of graph.

\subsection{Pyramid strategy}
Pyramid strategy is based on eigen-decomposition strategy. First, each graph is transformed into a bags-of-vectors using graph eigen-decomposition. Then, the vectors from bags-of-vectors are projected into multi-resolution histograms whose shape is like pyramid structure. After obtaining the pyramid structures of graphs, most of the existing pyramid graph kernels~\cite{2006Beyond,grauman2007the,Su2016A,Nikolentzos2017Matching} used histogram intersection to measure the similarity between two graphs. These graph kernels ignored the geometric structure information on each level of the pyramid. To address this problem, we project the graphs into pyramid structure and use optimal transport distance to measure the similarity between geometric structures on each level of the pyramid, seeing Figure~\ref{Pyramid_graph_OT}. Given a graph $G$, $f_{l}(G)$ denotes the pyramid embedding of graph $G$ at level $l$.

\subsection{Optimal transport distance on graphs}
By embedding the graphs into vector space, we could obtain the graph vectorial representations. Then, we can calculate the OT distance between pairwise graphs with definition 2. Before calculating OT distance, we need to choose appropriate ground distances to measure the similarity between each pair of nodes in graphs. If the node representations are discrete, we choose Hamming distance as ground distance. If the node representations are continuous, we choose Euclidean distance as ground distance. In order to facilitate the calculation of OT distance, we set $p=1$ in formula~(\ref{Wasserstein_distance}) to calculate $L^{1}$-Wasserstein distance. Given two graphs $G_1$ and $G_2$, $D^{f_E}_{OT}(G_1,G_2)$ denotes the OT distance based on graph eigen-decomposition embedding, $D^{f_l}_{OT}(G_1,G_2)$ denotes the OT distance based on pyramid embedding of graph $G$ at level $l$.
 \begin{equation}
    D^{f_E}_{OT}(G_1,G_2)=d_{OT}(f_{E}(G_1),f_{E}(G_2))
 \end{equation}

  \begin{equation}
    D^{f_l}_{OT}(G_1,G_2)=d_{OT}(f_{l}(G_1),f_{l}(G_2))
 \end{equation}

\begin{figure}
\centering
\includegraphics[width=8cm,height=5cm]{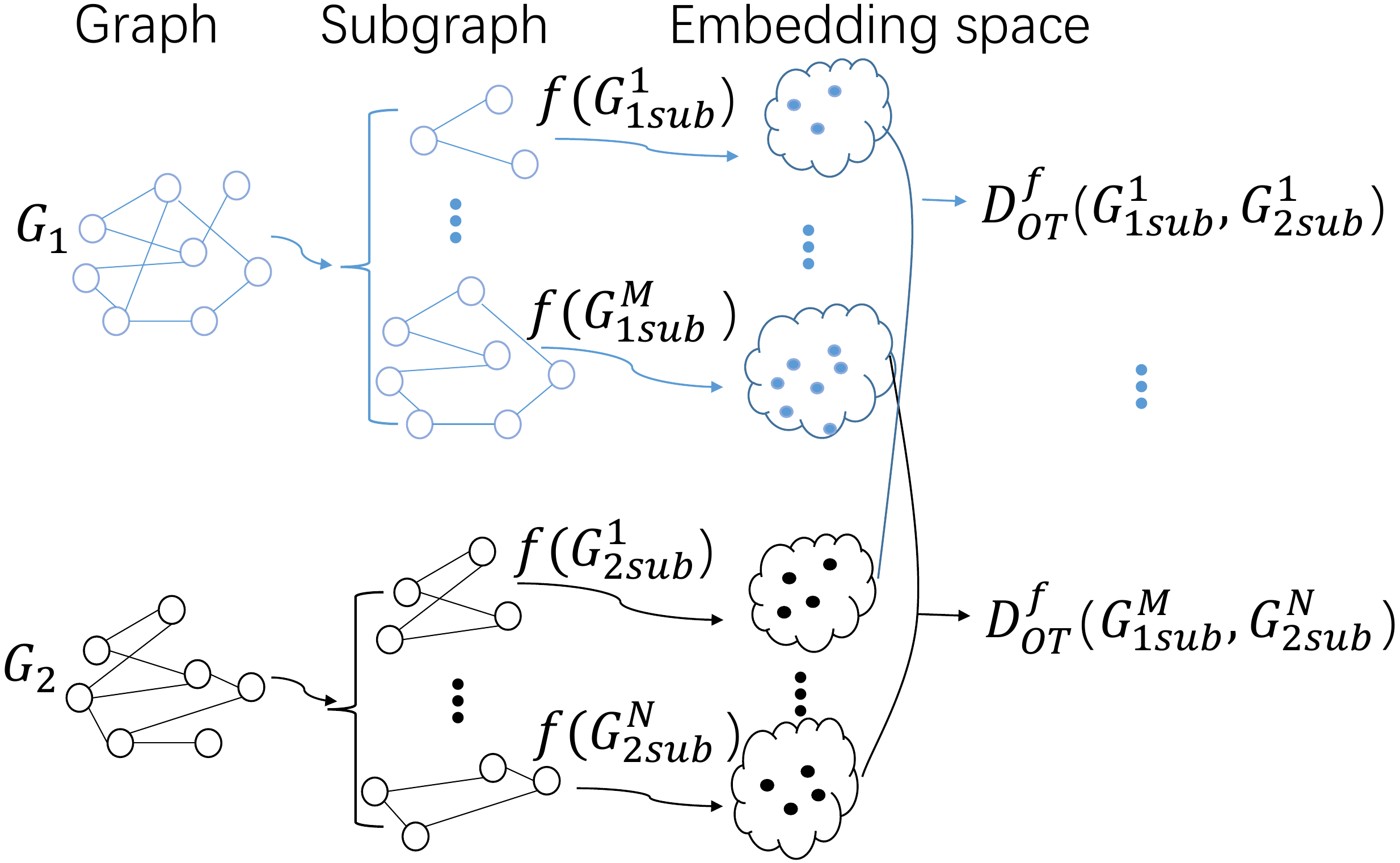}
\caption{Optimal transport distance based on subgraph structure}
\label{subgraph}
\end{figure}

\section{Optimal Transport Graph Kernels}
The optimal transport distances offer a more meaningful distance measurement for graph comparison tasks, it could capture the geometric information of local nodes. However, most of the existing graph kernels based on OT usually used optimal transport distance to measure the node label information, thereby ignoring node geometric structure information. Meantime, most of these kernels are usually not p.s.d. In this section, we take the OT distance into the graph hierarchical structure and sub-structure to design the related graph kernels, they are respectively called Pyramid graph kernel based on OT and Subgraph kernel based on OT. In order to obtain the p.s.d graph kernel, we adopt the learning method based on OT to learn the similarities between pairwise nodes, then design graph kernel to measure the similarity between two graphs. This kernel is called graph kernel learning by OT.

\subsection{Pyramid graph kernel based on OT}
Most of the existing pyramid graph kernel focused on histogram intersection, ignoring the geometric structure information on each level of the pyramid. For example, in the research of matching node embedding for graph similarity~\cite{Nikolentzos2017Matching}, they used the histogram intersections of pyramid structure to design graph kernel. They ignored the geometric structure information on each level of the pyramid. To address this problem, we propose pyramid graph kernel based on OT where we use the OT distance to measure the similarity between pairwise graphs on each level of the pyramid structure.

We embed each graph into the $d$ dimension vector space utilizing the related embedding methods such as eigen-decomposition strategy for obtaining the graph vector representation. In vector space, each node in graph corresponds to a point in $d$-dimension unit hypercub. We partition the  $d$-dimension unit hypercub into multiple cells and put these cells on the multi-layered pyramid structure (e.g., L levels). we put $2^l$ cells at level $l$ $(l\in L)$ of the pyramid structure. After all cells of each graph are put on the pyramid structure, the whole pyramid structure becomes multi-resolution histograms. At each level of the  pyramid structure, we use the OT distance to measure the similarity between two graphs. Optimal transport distance based on pyramid structure could be seen in Figure~\ref{Pyramid_graph_OT}. At last, we design the basic kernel based on OT distance at each level of the pyramid structure. We sum the basic kernel from each level of the pyramid structure and obtain the pyramid graph kernel based on OT.
\\ \\
\textbf{Definition 3.} (Pyramid graph kernel based on OT) For two graphs $G_1$ and $G_2$, $D^{f_l}_{OT}(G_1,G_2)$ is defined for two graphs $G_1$ and $G_2$ on their pyramid embedding at $l$ level, pyramid structure has $L$ levels, $l\in L$. We define the pyramid graph kernel based on OT (PG-OT kernel) between $G_1$ and $G_2$ as
\begin{equation}
   K_{PG}(G_1,G_2)=\sum^{L}_{l=1}k_{base}(D^{f_l}_{OT}(G_1,G_2))
\label{K_PG}
\end{equation}
where $k_{base}$ is a basic kernel function, it could be Linear kernel, Gaussian kernel, Laplacian kernel or other kernels.

\begin{algorithm}[t]
\caption{Calculate kernel $K_{OT}$} 
\hspace*{0.02in} {\bf Input:} 
Graph vectorial representation $Gvec$, max-iter, lamda, maxIter, C\\
     	lamda: 			parameter for sinkhorn update\\
		C:					parameter for regularization\\
		max-iter: 			max iteration of optimization\\
		maxIter:			max iteration of Sinkhorn Update\\
\hspace*{0.02in} {\bf Output:} 
Kernel matrix $K_{OT}$
\begin{algorithmic}[1]
\State Initialize: $u\leftarrow1$, $K_0\leftarrow Gvec^TGvec$,\\ $K\leftarrow K_0/max(K_0)$
\For{iter=1 to max-iter}
\State $M\leftarrow K_{ii}+K_{jj}-2K_{ij}$
    \State       $\mathrm{\textbf{K}}\leftarrow e^{-\lambda M}$
    \For{ i=1 to m}
    \For{ j=1 to m}
        \State $\textbf{u}\leftarrow\textbf{1}$
        \State $compt\leftarrow1$
        \While{compt< maxIter}
           \State $\textbf{u}\leftarrow Gvec_i/(\mathrm{\textbf{K}}(Gvec_j/\mathrm{\textbf{K}}^T\textbf{u}))$
          \State $compt\leftarrow compt + 1$
        \EndWhile
        \State $\textbf{v}\leftarrow Gvec_j/\mathrm{\textbf{K}}^T\textbf{u}$
        \State $P_K \leftarrow diag(\textbf{u})\mathrm{\textbf{K}}diag(\textbf{v})$
        \State $P\leftarrow P+P_K $

    \EndFor

    \EndFor
    \State $Diag\leftarrow diag(sum(P,1)+sum(P,0)-2diag(P))$
    \State $\nabla\leftarrow(-2P)-diag(diag(-2P))+Diag$
    \State $K\leftarrow K_0-(1/C)\nabla$
    \State $U,\sigma\leftarrow eig((K+K^T)/2)$
    \State $K\leftarrow U max(\sigma,0) U^T$
\EndFor

\State \Return K
\end{algorithmic}
\end{algorithm}

\subsection{Subgraph kernel based on OT}
The substructures in graphs uncover the local geometric information of graph. Most of the subgraph kernels or subgraph matching kernels utilized subgraph isomorphism or subgraph matchings to design graph kernels. These kernels usually focused on maximum common subgraph~\cite{Nikil2008Comparison,guohua2019couple} or graph attributes~\cite{kriege2012subgraph}, thereby ignoring the geometric information of subgraph. Meantime, finding isomorphism subgraph and common maximum subgraph are NP-hard~\cite{AM1979}. To address these problems, we propose subgraph kernel based on OT where we use the OT distance to measure the similarity between pairwise subgraphs. Optimal transport distance based on subgraph structure could be seen in Figure~\ref{subgraph}.
\\ \\
\textbf{Definition 4.} (Subgraph kernel based on OT) Given two graphs $G_1$ and $G_2$ which respectively has $M$ subgraphs and $N$ subgraphs, $G^{i}_{1sub}$ is the $i^{th}$ subgraph of graph $G_1$, $G^{j}_{2sub}$ is the $j^{th}$ subgraph of graph $G_2$. We define the subgraph kernel based on OT (SG-OT kernel) between $G_1$ and $G_2$ as
\begin{equation}
   K^E_{SG}(G_1,G_2)=\sum^{M}_{i=1}\sum^{N}_{j=1}k_{base}(D^{f_E}_{OT}(G^{i}_{1sub},G^{j}_{2sub}))
\label{K_E_SG}
\end{equation}
\begin{equation}
   K^P_{SG}(G_1,G_2)=\sum^{M}_{i=1}\sum^{N}_{j=1}K_{PG}(G^{i}_{1sub},G^{j}_{2sub})
\end{equation}
where $K^E_{SG}(G_1,G_2)$ is based on subgraph eigen-decomposition embedding and $K^P_{SG}(G_1,G_2)$ is based on subgraph pyramid embedding. The $k_{base}$ is a basic kernel function, it could be Linear kernel, Gaussian kernel, Laplacian kernel or other kernels.

\renewcommand\arraystretch{1.5}
\begin{table*}[t]
\centering
\begin{tabular}{lllllll}\hline \toprule
Method & MUTAG         & KKI           & NCI109        & PTC-MR        &  BZR          &COX2 \\\hline
SP                  &82.21$\pm$1.23 &51.78$\pm$3.41 &62.03$\pm$0.25 &56.14$\pm$0.38 &74.24$\pm$0.92 &70.18$\pm$0.63\\
GR                  &67.79$\pm$1.23 &47.37$\pm$0.12 &49.40$\pm$0.21 &58.14$\pm$0.78 &76.54$\pm$0.24 &77.42$\pm$0.32\\
OA                  &79.56$\pm$1.17 &49.81$\pm$2.21 &67.14$\pm$0.18 &56.28$\pm$0.36 &74.63$\pm$0.05 &72.58$\pm$0.53  \\
RW                  &76.89$\pm$0.87 &50.74$\pm$1.17 &56.28$\pm$0.34 &56.24$\pm$1.14 &77.92$\pm$1.12 &74.23$\pm$1.26  \\
PM                  &85.46$\pm$0.71 &52.23$\pm$2.24 &68.37$\pm$0.12 &58.66$\pm$0.62 &80.12$\pm$0.21 &77.23$\pm$1.24  \\
GH                  &71.81$\pm$0.34 &54.29$\pm$0.49 &56.57$\pm$0.12 &60.56$\pm$0.162 &76.12$\pm$0.03 &76.32$\pm$1.35  \\
GW                  &73.83$\pm$4.21 &53.16$\pm$3.4 &56.77$\pm$4.77 &61.11$\pm$6.7 &80.52$\pm$5.02 &77.92$\pm$8.73  \\
\textbf{RG-OT}      &70.77$\pm$0.32	&\textbf{58.64$\pm$0.82}	&58.64$\pm$0.02	&\textbf{66.74$\pm$0.09}	&78.9$\pm$0.02 &78.23$\pm$0.13\\
\textbf{PG-OT }      &\textbf{85.57$\pm$0.02}	&\textbf{61.25$\pm$1.69}	&64.68$\pm$0.08	&\textbf{61.4$\pm$0.01}	&\textbf{81.26$\pm$0.18} &\textbf{79.84$\pm$0.04}\\
\textbf{SG-OT}      &\textbf{88.72$\pm$0.39}	&\textbf{62.05$\pm$2.41}	&\textbf{69.57$\pm$0.06}	&\textbf{64.58$\pm$0.37}	&\textbf{85.65$\pm$0.13} &\textbf{80.15$\pm$0.32} \\\hline \toprule
\end{tabular}
\caption{Average classification accuracy (\%) on the graph datasets with no labels or attributes. SP is shortest path kernel. OA is optimal assignment similarity. GR is the graphlet kernel. RW is random walk kernel. PM is pyramid match kernel. GH is GraphHopper kernel. GW is Gromov-Wasserstein kernel. RG-OT is regularized graph kernel based on OT. PG-OT is pyramid graph kernel based on OT. SG-OT is subgraph kernel based on OT. }
\label{classification}
\end{table*}

\subsection{Regularized graph kernel based on OT}
In graph optimal transport problem, each graph is seen as a pile of earth, the nodes in graph is seen as the earth needing to be moved. The optimal transport distance between two graphs $G_1$ and $G_2$ is the minimum cost of transporting nodes (or earth) from one graph (a pile of earth) to the other graph (a pile of earth) when the cost matrix defined on the ground distance has been given. It follows that OT distance could be regarded as the distance of ground distance. The OT distance is not isometric and the metric space it induces depend on the chosen ground distance. Hence, obtaining a positive definite graph kernel based on OT distance is not necessarily possible.

OT distance between each pair of graphs is a minimum transportation cost, which is calculated on the ground distances of all nodes in the both of graphs. Here, we propose an idea whether the graph kernel could be obtained in the process of calculating the optimal transportation plans. We regard each graph as the object needing to be transported. In the process of transporting graphs, we design graph kernel based on OT to measure the similarity between each pair of graphs.
\\ \\
Given two graphs $G_i$ and $G_j$, $Gvec_i$ and $Gvec_j$ are the vectorial representation of these two graphs. The kernel $K_{ij}$ measuring the similarity between  $G_i$ and $G_j$ is defined as:
\\
\begin{equation}
    K_{ij}=K_{OT}(Gvec_i,Gvec_j)=\phi(Gvec_i)^T\phi(Gvec_j)
\label{Kernel_K}
\end{equation}
\\
The kernel $K_{ij}$ is the dot product of $Gvec_i$ and $Gvec_j$ in high dimension space. The cost matrix $M$ in formula (\ref{OT_distance}) is a ground metric which could be calculated by $M_{ij}=D^2_\phi(Gvec_i,Gvec_j)$.  Here, we find that there is a special relationship between ground metric $M$ and kernel $K_{OT}$. Its relationship could be represented as
\\
\begin{equation}
    M_{ij}=K_{ii}-2K_{ij}+K_{jj}
\end{equation}
\\
The kernel $K_{OT}$ as a regularizer is incorporated into formula (\ref{OT_distance}). The problem calculating minimum transportation cost in formula (\ref{OT_distance}) becomes the problem calculating the graph kernel $K_{OT}$. Hence, the graph kernel based on OT is also called regularized graph kernel based on OT (RG-OT kernel). Its calculation is defined as

\begin{equation}
\begin{split}
&min_{K_{OT}} \langle P,M\rangle+\frac{C}{2}\parallel K_{OT}-K_0\parallel^2_{F} \\
&s.t. ~~~~ K_{OT}\in \mathcal{S}_{+} \\
&~~~~~~~~M_{ij}=K_{ii}+K_{jj}-2K_{ij}
\end{split}
\label{Kernel_OT}
\end{equation}
where $C>0$ is a trade-off parameter, $\mathcal{S}_{+}$ is a set of positive semi-definite matrices. $K_0$ is a initialized matrix. The calculation of formula (\ref{Kernel_OT}) could refer to entropic regularization in OT~\cite{2013Sinkhorn}, the calculating process could be seen in \textbf{Algorithm 1}. Cuturi added the entropic regularization to the original problem in formula~(\ref{OT_distance}). The entropic regularization in OT is seen as an efficient approximation of the original problem, which is named as \emph{Sinkhorn distance}.
\\ \\
\textbf{Sinkhorn Distance}. Given a $d\times d$ cost matrix $M$ and two marginal distributions $r,c\in\sum_d$. The \emph{Sinkhorn distance} is defined as:
\begin{equation}
d^{\lambda}_{M}(r,c):=\langle P^\lambda,M\rangle
\label{Sinkhorndistance}
\end{equation}

\begin{equation}
P^\lambda=arg_{P\in U(r,c)}min\langle P,M\rangle-\frac{1}{\lambda}H(P)
\label{PSinkhorndistance}
\end{equation}
where $H(P)=-\sum^d_{i=1}\sum^d_{j=1}p_{ij}log (p_{ij})$ is the entropy of $P$ and $\lambda>0$ is entropic regularization coefficient.

The entropic regularization is very important for OT distance in formula~(\ref{OT_distance}). Because, it makes the objection function in OT distance become a strictly convex problem that could be solved through Sinkhorn's matrix scaling algorithm at a fast speed.

\section{Experiments}
In this section, we first introduce the datasets used in the experiments. Then, we introduce the related graph kernels which are compared with our algorithms and introduce the details about the experiment settings. Last, we perform our methods and the baselines on the given datasets and report the results.

\subsection{Datasets}
We perform the classification experiments on 6 widely used bioinformatics datasets, including MUTAG, KKI, NCI109, PTC-MR, BZR and COX2. These datasets could be downloaded from the online benchmark data sets for graph kernels~\cite{KKMMN2016}. MUTAG dataset consists of 188 graphs. KKI dataset consists of 83 graphs. NCI109 dataset consists of 4127 graphs. PTC-MR dataset consists of 344 graphs. BZR dataset consists of 405 graphs. COX2 dataset consists of 467 graphs. KKI is a brain network dataset which is constructed from the whole brain functional resonance image(fMRI) atlas. In graph, each vertex represents a region of interest(ROI), and each edge corresponds to correlations between two ROIs. KKI dataset is constructed for the task of Attention Deficit Hyperactivity Disorder (ADHD) classification.
MUTAG, PTC-MR and NCI109 are molecule compound datasets. BZR and COX2 are chemical compound dataset. The detailed introduction of the datasets could be seen in~\cite{KKMMN2016}.

\begin{figure*}
\centering
\includegraphics[width=14cm,height=4.5cm]{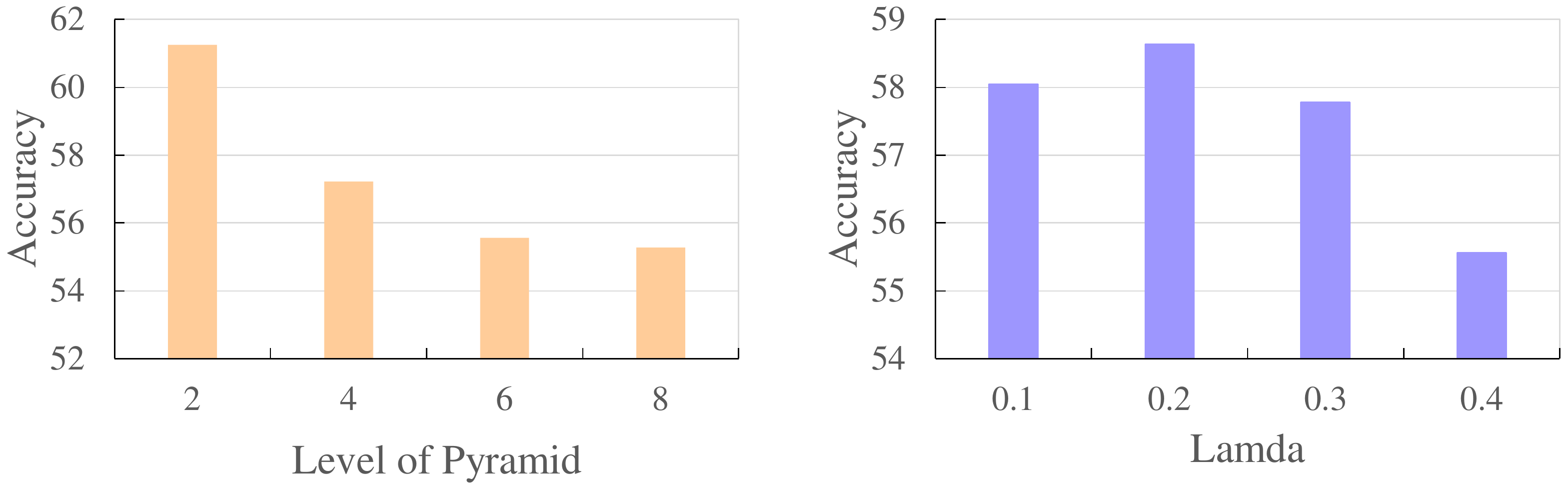}
\caption{Accuracies in different parameters in KKI dataset. Left is based on PG-OT kernel. Right is based on RG-OT kernel.}
\label{parameter}
\end{figure*}

\subsection{Baselines}
 We compare our kernel against several state-of-the-art graph kernels. In our experiments, we used the undirect and unlabeled graphs without attributes. Our baseline comparators include (1) shortest path kernel (SP)~\cite{Borgwardt2006Shortest} that calculates the number of shortest paths with equal length between  pairwise nodes, (2) optimal assignment similarity (OA)~\cite{johansson2015learning} that calculates the optimal assignment between the embeddings of the vectices of two graphs, (3) the graphlet kernel (GR)~\cite{sherashidze2009efficient} that counts common connected subgraphs of size 3, (4) random walk kernel (RW)~\cite{2010Graph} that calculate the number of common walks between two graphs, (5) pyramid match kernel (PM)~\cite{Nikolentzos2017Matching} that calculates the histogram intersections,
 (6) GraphHopper kernel (GH)~\cite{feragen2013scalable} that is a convolution kernel counting sub-path similarities that are truncated tree based graph kernels, (7) GW method~\cite{2019Optimal} that combines the Gromov-Wasserstein (GW) distance with shortest path.

\subsection{Experimental Setup}
We demonstrate our proposed graph kernels including pyramid graph kernel based on OT (PG-OT), subgraph kernel based on OT (SG-OT) and regularized graph kernel based on OT (RG-OT) which are introduced in the section of optimal transport graph kernels. In PG-OT kernel, we choose the level of the pyramid structure from \{2,4,6,8\}, seeing the left figure in Figure~\ref{parameter}. In the kernels of PG-OT and SG-OT, we use $L^1$-Wasserstein distance as our OT distance, and use the Laplacian kernel $K=e^{-\lambda D^f_{OT}}$ as the basic kernel $k_{base}$ in formula~(\ref{K_PG}) and~(\ref{K_E_SG}). In SG-OT, we use the kernel $K^E_{SG}(G_1,G_2)$ based on subgraph eigen-decomposition embedding as our subgraph kernel. There are all kinds of the subgraph decomposition methods. For example, subgraph decomposition based on shortest path~\cite{8272417} obtained the subgraph of each node. The subgraph of each node $v_i$ consisted of the node $v_i$ and the nodes with their shortest path length to node $v_i$. In random walk kernel, the walk setp is set as S=50. The parameter h in Weisfeiler-Lehman kernels is chosen from \{0,1,...,10\} by the cross-validation on the training set.

In regularized graph kernel based on OT, the max iteration of optimization is set as $maxiter= 100$. The parameter lamda for sinkhorn update is chosen from \{0.1, 0.2, 0.3, 0.4\}, seeing the right figure in Figure~\ref{parameter}. The parameter $C$ for regularization is set as $C=200$. The max iteration of Sinkhorn Update is set as $maxIter=200$.

We utilize the SVM as the classifier for the p.s.d graph kernels. If the graph kernel is not p.s.d, we use Krein SVM (KSVM)~\cite{2015Learning}, which is  specifically designed to process the indefinite kernels~\cite{2004Learning}, as the classifier. The tradeoff parameter $C$ in the SVM is selected from $\{10^{-3},10^{-2},\cdots,10^3\}$. The whole experiments are performed using the 10-fold cross-validation. We repeat each cross-validation split 10 times for each dataset and each method. We report the average accuracy in talbe~\ref{classification}

\subsection{Results}
The experiments demonstrate that our optimal transport graph kernels have the potential to make a contribution to the progress in real-world graph classification problems. The subgraph kernel based on OT outperform all the start-of-the-art graph kernels on all datasets except from PTC-MR dataset for unlabeled graphs. The classification accuracies obtained by pyramid graph kernel based on OT are better than those of the start-of-the-art graph kernels on MUTAG, KKI, PTC-MR, BZR and COX2 datasets. The regularized graph kernel based on OT achieve better classification accuracies than the start-of-the-art graph kernels on KKI and PTC-MR datasets. The classification accuracy achieved by the regularized graph kernel based on OT on PTC-MR dataset is the best.

\section{Conclusion}
In this paper, we respectively propose three graph kernels, they are subgraph kernel based on OT, pyramid graph kernel based on OT and regularized graph kernel based on OT. In subgraph kernel based on OT, OT distance is applied to measure the similarity between the subgraphs and this method could avoid calculating subgraph isomorphism. In pyramid graph kernel based on OT, we first embed the graphs into the pyramid structure where the hierarchies of the graphs are well considered. Then we apply the OT distance to measure the similarity between graphs in each hierarchy of the pyramid structure. In order to obtain the p.s.d graph kernel, we add the kernel regularization to the original OT distance and get the regularized graph kernel based on OT. We evaluate our methods on six datasets and compare them with several start-of-the-art graph kernels. The results demonstrate that our methods outperform these start-of-the-art graph kernels on most of the benchmark datasets.

\bigskip
\bibliographystyle{IEEEtran}
\bibliography{neurips_2020}

\end{document}